\documentclass[]{spie}  

\usepackage{amsmath,amsfonts,amssymb}
\usepackage{graphicx}
\usepackage{multirow}
\usepackage{amsmath}
\usepackage{makecell}
\usepackage{booktabs}
\usepackage[colorlinks=true, allcolors=blue]{hyperref}

\title{The Topology-Overlap Trade-Off in Retinal Arteriole-Venule Segmentation}

\author[a,b]{Ángel Víctor Juanco-Müller}
\author[b]{João F.C. Mota}
\author[a]{Keith A. Goatman}
\author[a]{Corné Hoogendoorn}
\affil[a]{Canon Medical Research Europe LTD, 2 Anderson Pl, EH6 5NP, Edinburgh, United Kingdom}
\affil[b]{Heriot-Watt University, Campus The Avenue, EH14 4AS, Edinburgh, United Kingdom}

\authorinfo{Further author information: (Send correspondence to Ángel Víctor Juanco-Müller)\\ E-mail: victor.juancomuller@mre.medical.canon}

\begin{document} 
\maketitle

\begin{abstract}
Retinal fundus images can be an invaluable diagnosis tool for screening epidemic diseases like hypertension or diabetes. And they become especially useful when the arterioles and venules they depict are clearly identified and annotated. However, manual annotation of these vessels is extremely time demanding and taxing, which calls for automatic segmentation. Although convolutional neural networks can achieve high overlap between predictions and expert annotations, they often fail to produce topologically correct predictions of tubular structures. This situation is exacerbated by the bifurcation versus crossing ambiguity which causes classification mistakes. This paper shows that including a topology preserving term in the loss function improves the continuity of the segmented vessels, although at the expense of artery-vein misclassification and overall lower overlap metrics. However, we show that by including an orientation score guided convolutional module, based on the anisotropic single sided cake wavelet, we reduce such misclassification and further increase the topology correctness of the results. We evaluate our model on public datasets with conveniently chosen metrics to assess both overlap and topology correctness, showing that our model is able to produce results on par with state-of-the-art from the point of view of overlap, while increasing topological accuracy.

\end{abstract}

\section{Introduction}
\label{sec:intro}  
The fundus of the eye provides a window to the human microvasculature and can be imaged easily from 2D colour photographs. Whenever an accurate arteriole-venule (AV) segmentation of the retinal fundus image is available, it becomes possible to compute important biomarkers, like the arteriole to venule ratio, which can be used to diagnose conditions such as hypertension and diabetes \cite{Stokoe1996,Vazquez2013}. However, the required blood vessel delineation and classification are intensive and laborious tasks, which may create bottlenecks in clinicians workloads. As a result, automatic arteriole-venule (AV) segmentation models are of great interest. 

A variety of automatic image segmentation algorithms have been proposed, including convolutional neural networks (CNN) which achieve state-of-the-art results in segmentation tasks \cite{Long2015}. However, such networks are often assessed using overlap-specific metrics such as the Dice and Jaccard scores \cite{AbdelAziz2015}, which do not take into account desirable features for tree-like tubular segmentation. Therefore, we consider relevant measures, which we term topology metrics \cite{hemmelings_code,atm_repo}, and use them to evaluate the proposed solution.

To include tubular priors in a convolutional neural network, we use a topology loss and an orientation score guided module. Whereas the former focuses the loss on foreground pixels that are closer to the centerline rather than to the boundaries, which reduces discontinuities in the predicted vessels, the latter leverages orientation-sensitive filters implemented with cake wavelets\cite{Dutis2007,Feliciano2016,Abbasi-Sureshjani2016}, which help to disambiguate crossing regions. We evaluate our model in the DRIVE-RITE \cite{Staal2004,Qureshi2013} and LES-AV \cite{Orlando2018} public datasets and find that it performs competitively with the state-of-the-art of AV segmentation\cite{Hemelings2019,Galdran2019,Luo2022} in terms of overlap metrics while consistently achieving superior topology metrics.

The paper is structured as follows. We review related work in \S\ref{sec:related_work}, describe the network architecture, loss function and experimental details in \S\ref{sec:Materials & methods}, present the results in \S\ref{sec:Results}, and discuss and conclude the paper in \S\ref{sec:Discussion}.

\section{Related Work}
\label{sec:related_work}
Researchers have tackled retinal AV segmentation with a variety of methods, ranging from specifically crafted traditional algorithms to powerful although less interpretable deep learning models. The problem, however, remains challenging. Because fundus images provide a 2D projection of the 3D arteriole and venule trees\cite{Estrada2015}, differentiating between bifurcations and intersections in these images becomes an NP-hard problem \cite{Estrada2014}. 

Traditional methods for AV segmentation generally consist of two stages: vessel segmentation and classification. For the former task, filters suited for tubular detection, such as the \emph{Frangi} and \emph{Hessian} kernels \cite{Frangi1998,Otsu1998}, are used together with morphological operations \cite{Mendoza2006} to obtain vessel segmentations. For the latter, machine learning classifiers are employed to classify vessel segments as arterioles or venules \cite{Akbar2018,Fraz2014}. 

Because oxygenated haemoglobin absorbs less light between the 600 and 800 nm wavelength window \cite{Huang2018}, arteries reflect more light and hence are brighter than veins. Thus, in theory colour intensity could be used for classification. However, inhomogeneous illumination conditions are common in retinal fundus images, requiring illumination preprocessing stages \cite{Galdran2018}. Furthermore, intensities measured at the smaller vessel segments may be noisier and less useful for classification \cite{Joshi2014}.

Some approaches connect the different segments to build a graph where the nodes are called \emph{junctions points}, which can be of three types: endpoints, bifurcations and crossings. The task then consists of identifying subgraphs that do not contain crossing nodes and correspond to the true vessel trees. Dashtbozorg et al.\cite{Dashtbozorg2013} approaches this task with a graph analysis technique, and Joshi et al.\cite{Joshi2014} with a Dijkstra graph search. However, these approaches are limited by the the accuracy of the vessel segmentation used to build the graph. 

To mitigate this issue, some methods aim to extract and classify the junction points directly from the images. For this task, anisotropic wavelets, such as the single sided cake wavelet \cite{Dutis2007}, have been used to extract a set of orientation responses. These features, when combined with other vessel segmentation methods, have proved to be helpful for junction point classification \cite{Abbasi-Sureshjani2016,Feliciano2016}.

Nevertheless, the presence of pathology can alter the eye appearance in the fundus images \cite{Stokoe1996,Vazquez2013} and therefore decrease the performance of the above mentioned algorithms. For example, hypertension increases vessel sinuosity and diabetic retinopathy produces lesions that can change colour and intensity of the vessels \cite{Wang22}. These issues may be rectified by deep learning models trained in sufficiently large datasets with enough variability of cases. For this reason, solutions based on convolutional neural networks (CNNs) \cite{Long2015,Ronneberger2015,Insensee2019} have become popular among researchers.

To learn the topology of the vessels, Hemelings et al. \cite{Hemelings2019} increase the receptive field in CNNs with dilated convolutions. To capture both thicker and thinner vessels, Kang et al. \cite{Kang2020} combine different kernel sizes \cite{Szegedy2015}. Galdran et al. \cite{Galdran2019} consider both crossings and uncertain vessels to be of the same class. More conveniently, Morano et al. \cite{Morano2021} build a ground truth image whose channels correspond to continuous arteriole, venule and vessels, and minimize the binary cross entropy separately for each class, which demonstrated improved performance, and it is the strategy we adopt here. Recently, Luo et al. \cite{Luo2022} proposed an segmentation network followed by a conditional generative adversarial model for AV refinement, which also inspires the two-stage nature of the method hereby presented. Nevertheless, all these methods still fail to preserve the continuity of segmented arterioles and veins.

\section{Materials \& methods}
\label{sec:Materials & methods}
Here we describe the preprocessing stage in \S\ref{subsec:Preprocessing}, the proposed model architecture in \S\ref{subsec:Network architecture}, the loss function with topology terms in \S\ref{subsec:Topology preserving losses} and give details of the experimental set-up in \S\ref{subsec:Experimental Set-up}.

\subsection{Preprocessing}
\label{subsec:Preprocessing}
\begin{figure}
	\includegraphics[width=\textwidth]{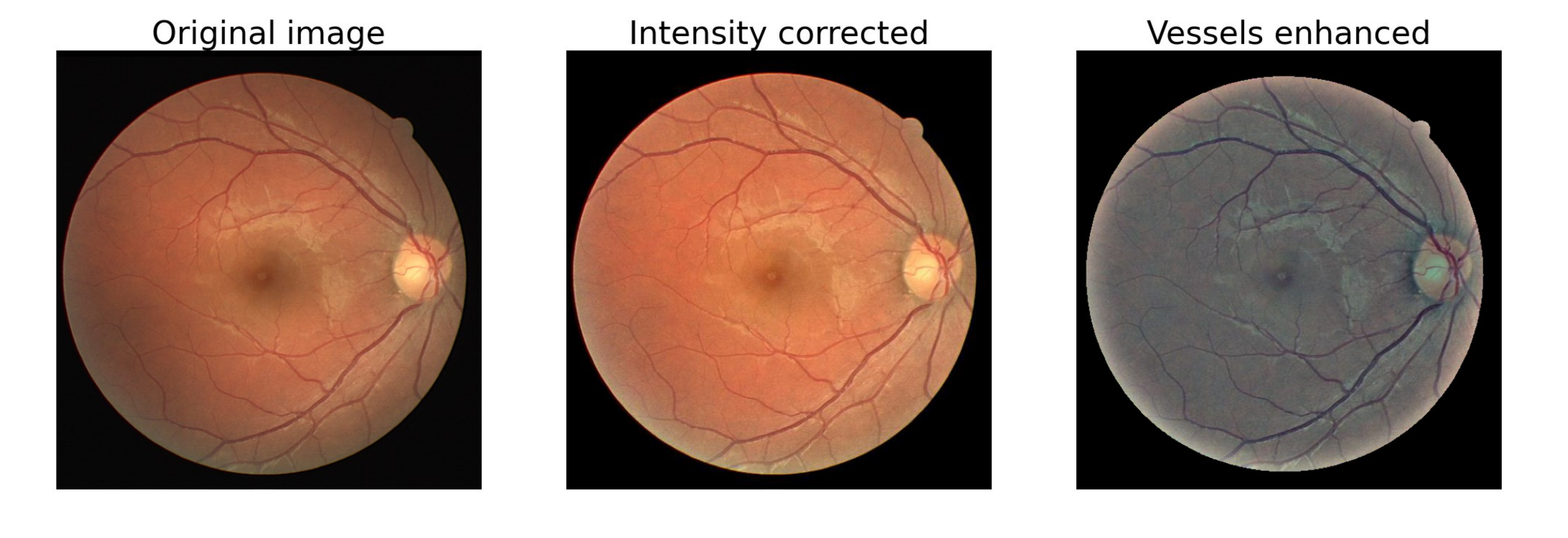}
	\caption{Preprocessing of sample \emph{37\_training} DRIVE-RITE dataset \cite{Staal2004,Qureshi2013}. Left: original image. Middle: image after illumination correction \cite{Kaiming2009,Galdran2018}. Right: image after vessel enhancement \cite{Morano2021}.} \label{fig:preprocessing}
\end{figure}

Before feeding the images to the proposed model, we perform two preprocessing stages. First, to mitigate the heterogeneous illumination of fundus images, we apply an intensity homogenization procedure\cite{Galdran2018} based in the dark channel prior \cite{Kaiming2009}. Next, we enhance the vessels by subtracting to the image its Gaussian filter response. This has the effect of a high pass filter and separates the modes of foreground and background intensity distributions \cite{Morano2021}. Figure \ref{fig:preprocessing} shows the different stages for a sample of the RITE dataset \cite{Staal2004,Qureshi2013} affected by inhomogeneous illumination.

\subsection{Network architecture}
\label{subsec:Network architecture}
\begin{figure}
	\begin{center}
		\begin{tabular}{c}
			\includegraphics[width=\textwidth]{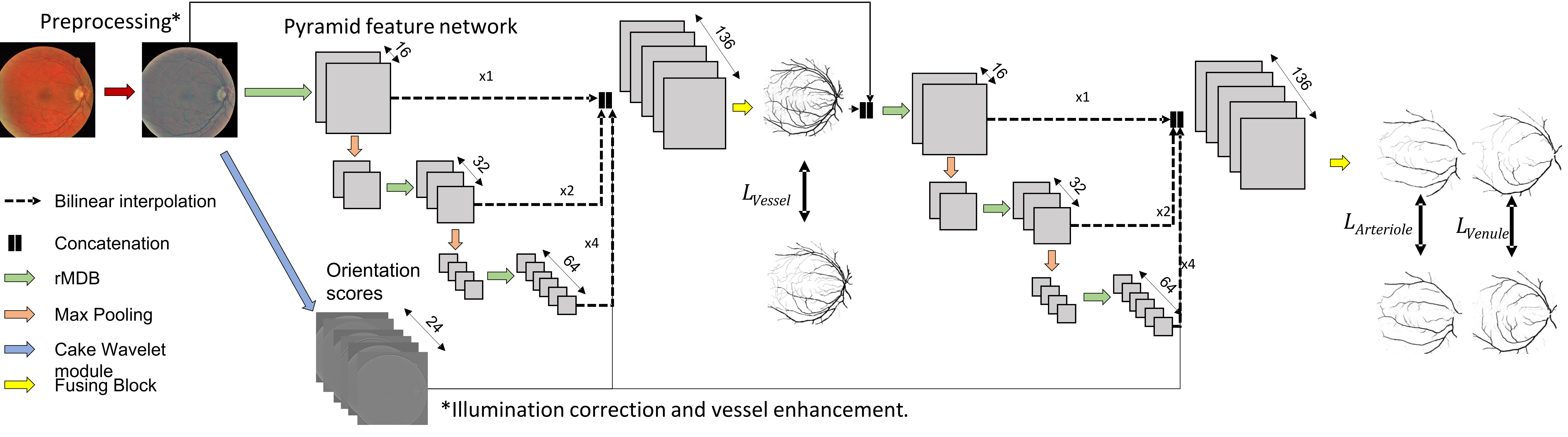}
			\\\includegraphics[width=\textwidth]{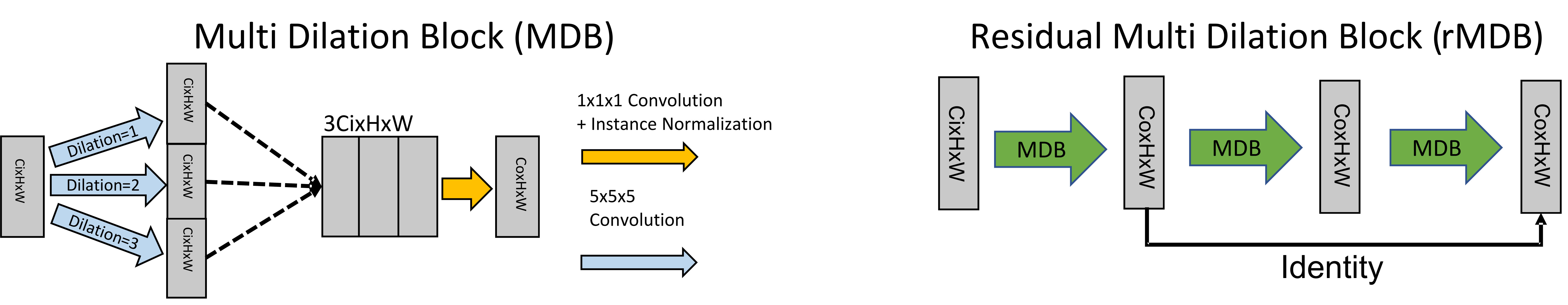}
			\\\includegraphics[width=\textwidth]{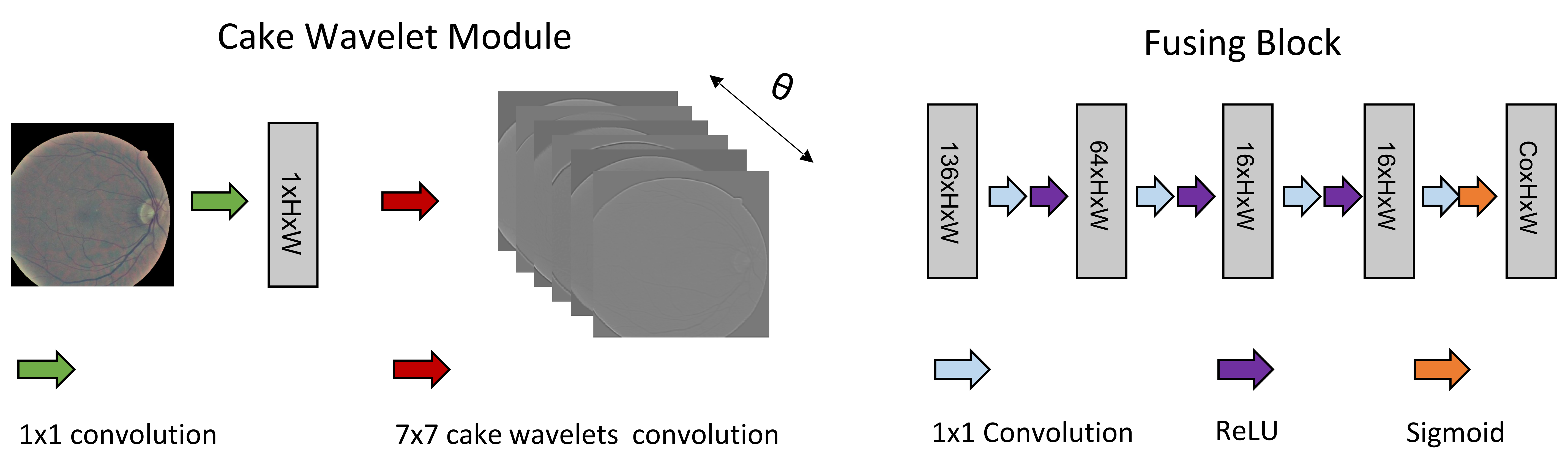}
		\end{tabular}
	\end{center}
	\caption{Top: proposed pipeline; from left tor right: preprocessing, neural network processing and loss calculation. Middle, from left to right: multi-dilation block (MBD) which combines 5x5 convolutions at different dilation and its residual implementation (rMBD). Bottom, from left to right: Cake wavelet module, with a 7x7 convolution initialized with the negative real part of cake wavelets, and the fusing block, a combination of 1x1 convolutions and ReLU activations except at the final layer, where sigmoid is used.} \label{fig:model_and_blocks}
\end{figure}

The proposed neural network architecture for AV segmentation, depicted in the top of Fig. \ref{fig:model_and_blocks}, broadly consists of two feature pyramid networks, explained in \S\ref{subsubsec:Feature pyramid network}, and the cake wavelet module, described in \S\ref{subsubsec:Cake wavelet module}. The building blocks of the feature pyramid networks are residual multi-dilation convolutional layers, as explained in \S\ref{subsubsec:Residual multi-dilation block}. Two fusing blocks, presented in \S\ref{subsubsec:Fusing block}, process the output features of the feature pyramid network and cake wavelet module to produce the output segmentation results. The first feature pyramid block is responsible for segmenting the vessels, whereas the second one, conditioned on the vessel probability and input image and leveraging the orientation sensitivity of the cake wavelets, produces the AV segmentation.

\subsubsection{Feature pyramid network}
\label{subsubsec:Feature pyramid network}
The feature pyramid network has three levels of depth. At the first level, the output feature map contains 16 channels, and this number is doubled at every block. The resulting feature maps are both downscaled by a factor of 2 with max-pooling for downstream processing in the feature pyramid network and resized to the original image size for input to the fusing block. 

\subsubsection{Residual multi-dilation block}
\label{subsubsec:Residual multi-dilation block}
The multi-dilation block applies convolutions of kernel size 5x5 with dilation values of 1, 2 and 3 in parallel. The output responses are then concatenated and processed by a 1x1 convolution and instance normalization layers\cite{Ulyanov2016} to produce the output feature map, as shown in the middle left of Fig. \ref{fig:model_and_blocks}. This operation is applied three consecutive times with a residual connection\cite{He2016} between the second and the third output, as depicted in the middle right of Fig. \ref{fig:model_and_blocks}.

\subsubsection{Cake wavelet module}
\label{subsubsec:Cake wavelet module}
\begin{figure}
	\includegraphics[width=\textwidth]{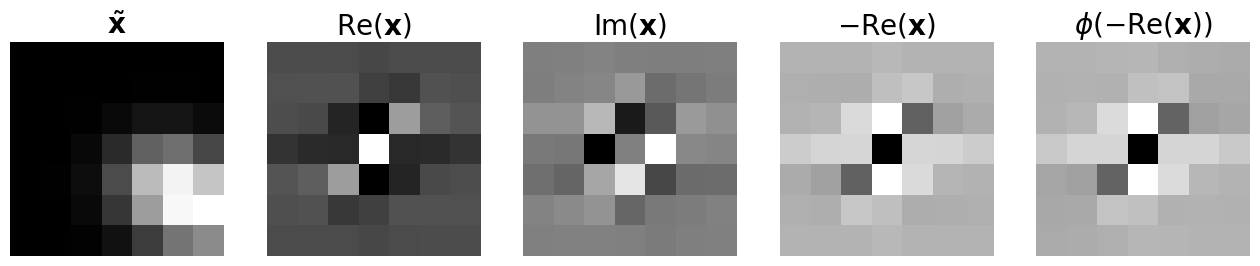}
	\caption{\label{fig:wave} Cake wavelets for $\theta=45º$. From left to right: oriented shape in frequency domain $\mathbf{\tilde{x}}$, real and imaginary part of the inverse Fourier transform $\mathbf{x}$, negative real part of $\mathbf{x}$ before and after training of the CNN model $\phi$.}
\end{figure}

The single-sided cake wavelets \cite{Dutis2007} are defined as the inverse 2D Fourier transform, $\mathbf{x}$, of an oriented shape in frequency domain, $\mathbf{\tilde{x}}$. With an appropriate window size, the real and imaginary parts of these filters approximate an \emph{orientation detector} perpendicular to the orientation in frequency domain (see Fig. \ref{fig:wave}, left).

The cake wavelet module first maps the RGB fundus image to grayscale with a 1x1 convolution and then convolves it with 7x7 convolutions initialized with the negative real part of the single side cake wavelets filters for 24 values of an orientation angle $\theta\in\{0, 2\pi\}$. An schematic of the block is shown in the bottom left of Fig. \ref{fig:model_and_blocks}.

Although we initialize the cake wavelet module with the cake wavelet filters, we let the network learn them during training. It turns out that after training the values of the filters remain practically unchanged, which shows the suitability of the cake wavelet for the task at hand (see Fig. \ref{fig:wave}, right). 

\subsubsection{Fusing block}
\label{subsubsec:Fusing block}
The fusing block takes the concatenation of outputs at the feature pyramid module at different resolution levels and the orientation scores produced by the cake wavelet module. These are all concatenated resulting in a feature map with 136 channels. A series of 1x1 convolution layers and ReLU activations reduce the channel dimension from 136 to 64, from 64 to 16, and finally from 16 to the output number of segmentation channels.

\subsection{Loss Function}
\label{subsec:Topology preserving losses}
We optimize a combination of topology and overlap losses. We set the Dice loss function $L_{\text{Dice}}$ as the overlap term. To improve the connectivity of the predicted vessels, a topology term enhances the vessel centerline, that is the inner continuous line equidistant to both vessel walls. We adopt the centerline-Dice $L_{\text{cl-Dice}}$ proposed by Suprosanna and Shit \cite{Shit2021}, described in \S\ref{subsubsec:Centerline Dice}. 

Furthermore, we include \emph{mixed terms}, which are the Mean Square Error $L_{\text{MSE}}$ and Binary Cross Entropy $L_{\text{BCE}}$ losses weighed by a centerline cost-map $\alpha$ according to the policy described in \S\ref{subsubsec:Centerline Cost Map}. Thus, the network is trained to minimize the following loss
\begin{equation}
	\label{eq:overall_loss}
	L_{\text{class}} = \overbrace{\lambda_1 L_{\text{Dice}}}^{\text{Overlap term}} + \overbrace{\lambda_2 L_{\text{cl-Dice}}}^{\text{Topology term}} + \overbrace{\alpha \left(\lambda_3 L_{\text{MSE}} + \lambda_4 L_{\text{BCE}} \right)}^{\text{Mixed terms}},
\end{equation}
where $\text{class} \in \{\text{arteriole}, \text{venule}, \text{vessels}\}$, and $\lambda_1$, $\lambda_2$, $\lambda_3$ and $\lambda_4$ are constants experimentally set to $1$, $0.5$, $0.5$ $0.5$ respectively.

\subsubsection{Centerline Dice}
\label{subsubsec:Centerline Dice}
\begin{figure}
	\begin{center}
		\includegraphics[width=\textwidth]{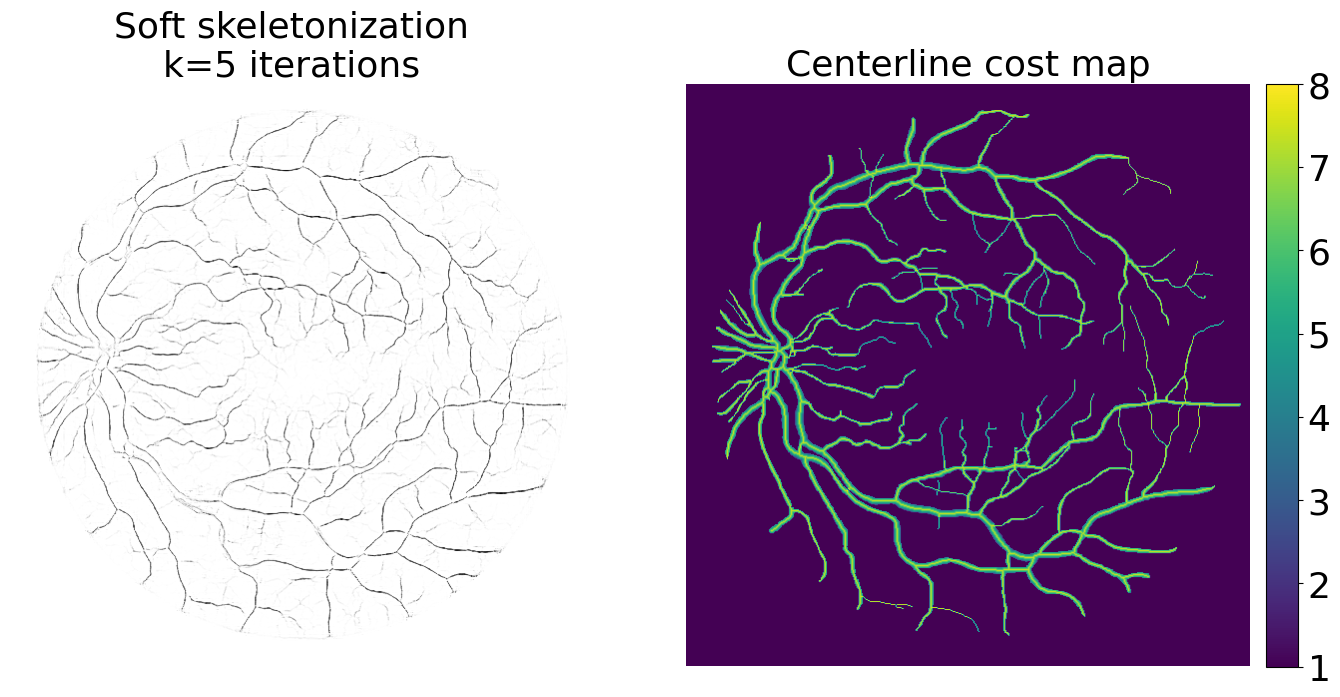}
		\caption{Video 2. Left: iterative soft skeletonization. Setting k=5 was sufficient to achieve good results. Right: centerline geodesic distance transform to weigh foreground pixels in the loss computation.} \label{fig:soft_skel}
	\end{center}
\end{figure}

The centerline Dice \cite{Shit2021} applies max and min pool functions to a vessel likelihood map for a number of iterations $k$ to retrieve the centerline or skeleton. Because min and max pool are differentiable functions, the resulting skeleton probability can be used to compute the Dice loss with respect to the skeleton ground truth and backpropagate gradients to directly improve the structural connectivity of the outputs. Fig. \ref{fig:soft_skel} left, shows how the number of iterations $k$ relates with the quality of the centerline; in our work setting $k=5$ was sufficient to achieve good results without considerably increasing complexity.

\subsubsection{Centerline Cost Map}
\label{subsubsec:Centerline Cost Map}
To balance the scale of the gradients a \emph{cost-map} $\alpha$ is used to weight pixels in the centerline more than those on the vessel walls. The cost-map $\alpha$ is defined as the geodesic distance transform with respect to the ground truth centerline restricted to the ground truth vessels (Fig. \ref{fig:soft_skel}, right). However, to allow the network first learn the overall vessel shape, and then correct for connectivity mistakes, $\alpha$ weights all pixels equally until the vessel predictions reach a minimum Dice score ($\sim0.6$), and only after that $\alpha$ is set to the described centerline distance transform.

\subsection{Experimental Set-up}
\label{subsec:Experimental Set-up}
This section explains our implementation details, starting with data augmentation in \S\ref{subsubsec:Data Augmentation}, and then presenting the used hardware, software and other training details in \S\ref{subsubsec:Soft and Hard}.

\subsubsection{Data Augmentation}
\label{subsubsec:Data Augmentation}
We trained our network on 17 images from the DRIVE-RITE training set \cite{Staal2004,Qureshi2013} and withheld the three remaining samples to validate the model after each epoch. To enlarge the number of training samples we applied a variety of augmentation techniques: rotations, flips, intensity shifts and elastic deformations were used to build eight augmented patches from each training sample. To better capture vessel structure, we set a large patch size ($512\times512$; the full images are $584 \times 565$). At inference we used a $512\times512$ sliding window with a $10\%$ overlap.

\subsubsection{Software, Hardware and Training details}
\label{subsubsec:Soft and Hard}
The preprocessing stages were implemented with the Scikit-Image library \cite{vanderWalt2014}. For the neural networks, we used the Pytorch Deep Learning framework \cite{pytorch} with the ADAM optimizer \cite{Kingma2014}. The MONAI library \cite{Monai} was used for data augmentation and loading. For the geodesic distance transforms, we used the scikit-fmm package \cite{scikit-fmm}.

We tracked a combination of two overlap (Dice score and precision) and two topology (tree detection rate and branch detection rate) metrics \cite{atm_repo}. We assumed convergence whenever the tracked metric did not improve for more than 10 epochs.

The experiments were run using a NVIDIA Tesla V100 with 32 GB of RAM. The average memory and time by training epoch was 5 GB and 3 minutes with batch size of 1. The learning rate, initially set to 0.001, was reduced by a factor of 0.1 whenever this metric did not improve for 5 epochs. Total training was completed between 2 and 3 hours.

\section{Results}
\label{sec:Results}
This section presents the AV segmentation results. We give a first evaluation with the receiving operator characteristic (ROC) curve in \S\ref{subsec:ROC}. We then introduce the topology sensitive metrics in \S\ref{subsec:metrics}. Next, we present the results in the DRIVE-RITE \cite{Staal2004,Qureshi2013} dataset in \S\ref{subsec:avRite}, and finally the results on the LES-AV dataset \cite{Orlando2018} in \S\ref{subsec:avLesAV}.

\subsection{ROC Analysis}
\label{subsec:ROC}
\begin{figure}
	\begin{center}
		\includegraphics[scale=0.5]{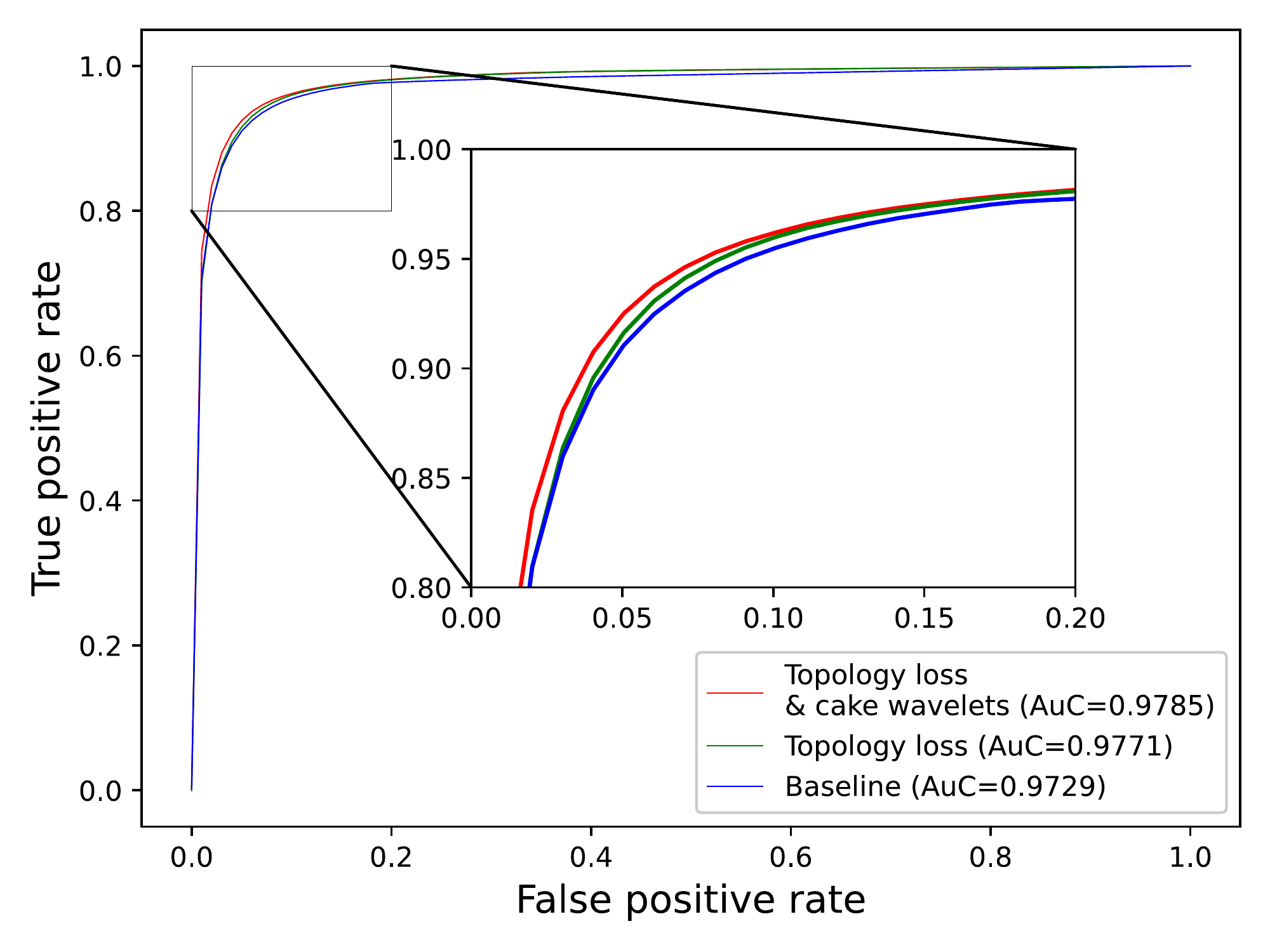}
		\caption{ROC curves computed on the 20 test images of the DRIVE-RITE dataset \cite{Staal2004,Qureshi2013} for the three proposed models. As it can be seen, the version with both topology loss and cake wavelets results in the curve enclosing the highest area.} \label{fig:roc_rite}
	\end{center}	
\end{figure}

Fig. \ref{fig:roc_rite} presents the ROC of the baseline model, and the two proposed modifications, computed on the 20 test images of the DRIVE-RITE dataset \cite{Staal2004,Qureshi2013}. It shows that all three models achieve high areas under the curve (AuCs), with a slight improvement for the model with both topology loss and Cake wavelets. Given the high AuC for all models, we decided to use a threshold of $0.5$ binarize the model outputs.
\subsection{Overlap, Mixed \& Topology Metrics}
\label{subsec:metrics}
To assess both overlap and topology correctness we devise specific metrics and organise them in three groups: \emph{overlap}, \emph{mixed}, and \emph{topology} ones.

The \emph{overlap} and \emph{mixed} metrics are derived from the evaluation code released by Hemelings et al. \cite{Hemelings2019}, and they calculate the F1 and accuracy of the AV classification on all pixels (\emph{overlap}) and only on the ground truth centerline pixels (\emph{mixed} metrics). The \emph{topology} metrics are the vessel detection rate, also from Hemelings et al.\cite{Hemelings2019}, and the tree length and branch detection rates, from the Airway Tree Modelling 2022 (ATM22) challenge \cite{atm_repo}.

The vessel detection rate is the prediction accuracy computed only on ground truth vessel pixels. The tree length detection rate is the sensitivity of the prediction with respect to the ground truth centerline. Finally, the branch detection rate uses an image where ground truth vessel segments are uniquely labelled to determine how many of them are present in the predicted segmentation.

Both the \emph{topology} and \emph{mixed} metrics are more difficult to optimize because they provide local measurements, constrained to pixels in certain regions of the images, and therefore are more sensitive to the predicted vessel topology than overlap measurements evaluated in all pixels.

\subsection{AV Segmentation Results on DRIVE-RITE}
\label{subsec:avRite}
We compared the baseline model against the version with topology loss and the one with topology loss and Cake wavelet module. For each combination, we trained 5 models and used an ensemble of them to produce the final output. Table \ref{tab:results_rite} reports results using the obtained segmentations and the masks shared by Hemelings et al. \cite{Hemelings2019}.
\begin{table}[t]
	\caption{\label{tab:results_rite}AV classification results in the RITE test set \cite{Staal2004,Qureshi2013}. Although our method achieves slightly worse results than Hemelings et al \cite{Hemelings2019} at the overlap level, it performs best in the topology and mixed metrics. Best figures are in bold.}
	\begin{tabular}{llllllll}\toprule
		\multirow{2}{*}{}                        & \multicolumn{2}{l}{\thead[c]{Overlap metrics\\ (evaluated on all   pixels)}} & \multicolumn{2}{l}{\thead[c]{Mixed Metrics \\ (evaluated on centerline pixels)}} & \multicolumn{3}{l}{\thead[c]{Topology metrics \\ (detection   rates)}} 
		
		\\ \cmidrule(lr){2-3} \cmidrule(lr){4-5} \cmidrule(lr){6-8}
		& F1                            & Accuracy                        & F1                             & Accuracy                          & Branches          & Tree length         & Vessel         \\\midrule
		Baseline                                 & 95.85                         & 95.98                           & 68.75                         & 55.88                             & 43.73             & 56.73               & 66.63          \\
		Hemelings et al. \cite{Hemelings2019}                 & \textbf{96.71}                         & \textbf{96.78}                           & 76.75                          & 66.65                             & 50.97             & 66.61               & 73.06          \\
		\thead[l]{Baseline +\\ topology loss}                 & 95.31                         & 94.97                           & 76.32                          & 68.99                             & 58.27             & 72.13               & 78.14          \\
		\thead[l]{Baseline +\\ topology loss +\\ cake wavelets} & 95.87                         & 95.66                           & \textbf{76.84}                         & \textbf{69.85}                             & \textbf{59.96}             & \textbf{72.93}               & \textbf{77.85}      \\\bottomrule   
	\end{tabular}
\end{table}
\begin{table}
	\caption{\label{tab:results_lesav} Results in the LES-AV dataset \cite{Orlando2018} of the model trained on RITE \cite{Staal2004,Qureshi2013} compared to other researcher's methods. Best figures are in bold.}
	\begin{center}
		\begin{tabular}{llll}
			\toprule Method & Accuracy & F1    & Specificity \\\midrule
			Galdran et al. (2019) \cite{Galdran2019} & 86.00    & 86.00 & 85.00       \\
			Luo et al. (2022) \cite{Luo2022} & 95.10    & 96.10 & 94.10       \\
			Ours                   & \textbf{96.12}    & \textbf{96.40} & \textbf{97.20}      \\\bottomrule
		\end{tabular}
	\end{center}
\end{table}

Although our model performs slightly worse than Hemelings et al \cite{Hemelings2019} at the all-pixel level, it outperforms their method at the centerline level and in terms of the topology metrics. The lower overlap quality is a side effect of optimizing the models for topology correctness, which is mitigated with the cake wavelet module. Further design strategies are needed to balance the topology-overlap trade-off.

\begin{figure}
	\includegraphics[width=\textwidth]{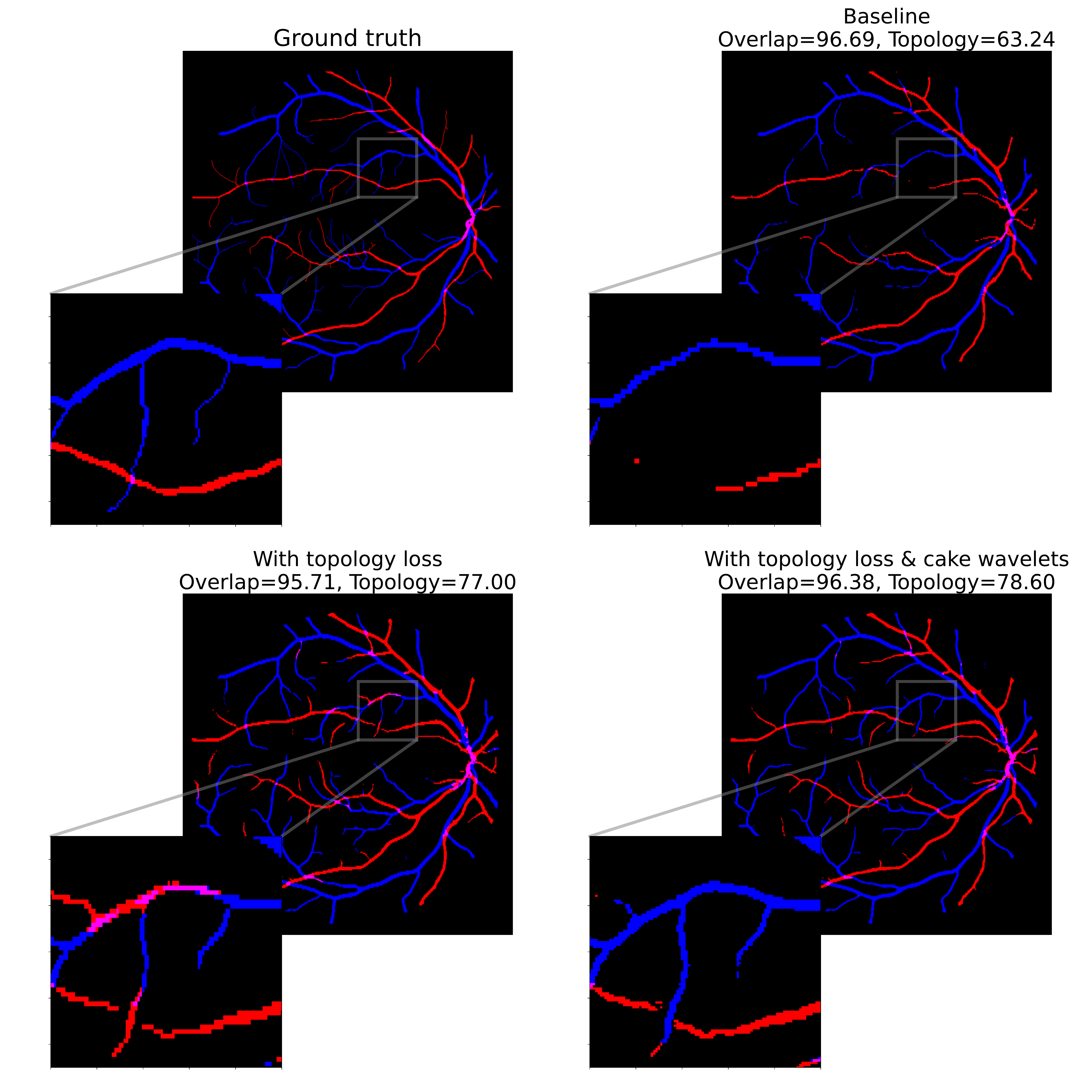}
	\caption{Top left: Ground truth of sample \emph{test07} of the DRIVE-RITE dataset \cite{Staal2004,Qureshi2013}. Top right: the prediction of our baseline model, which achieves high overlap at the cost of continuity mistakes. Bottom left: the result of model with topology loss, which produces continuous vessels although with misclassified regions. Bottom right: the output of the model trained with both topology loss and orientation scores, which produces continuous and correctly classified vessels. Overlap is the average of F1 and accuracy as in \cite{Hemelings2019}, and topology is the average of tree, branch \cite{atm_repo} and vessel detection rates \cite{Hemelings2019}.} \label{fig:rite_vis}
\end{figure}

Finally, to give a qualitative interpretation of the results, Fig. \ref{fig:rite_vis} illustrates the effect of the proposed modifications to the baseline model. As it can be seen, the model trained with the topology loss function produces continuous vessels, however at the cost of misclassifying arterioles and venules pixels within them. Including the Cake Wavelet module in addition to the topology loss mitigates this issue.

\subsection{AV Segmentation Results on LES-AV}
\label{subsec:avLesAV}
To explore the generalizability of the proposed model to unseen images we apply it to the LES-AV dataset \cite{Orlando2018}, which contains fundus images and AV segmentations of size $1620\times1444$. We resized the samples in LES-AV \cite{Orlando2018} with bilinear interpolation to the patch size used for training, that is $512\times512$, and retrieve the binary reference segmentations by thresholding at $0.5$.

Table \ref{tab:results_lesav} presents AV classification results compared to other state of the art approaches in the LES-AV dataset \cite{Orlando2018}. Because topology metrics are not reported in these papers, we limit our assessment to overlap (pixel-wise) measures. We again used the code released by Hemelings et al. \cite{Hemelings2019} to compute the metrics, and reported results in other authors' papers.  In spite of being trained with only 17 images from the RITE dataset \cite{Staal2004,Qureshi2013}, our method performs competitively with the LES-AV dataset \cite{Orlando2018}.

\section{Discussion \& Conclusion}
\label{sec:Discussion}
We have proposed a method to tackle the overlap-topology trade-off in arteriole-venule segmentation in fundus images. Our algorithm, however, is general: it is applicable to any task involving segmentation of complex tubular structures. For this reason, we adopted metrics used for pulmonary airway segmentation \cite{atm_repo}, and used them to validate the effect of the topology loss terms in our loss function. 

We found that only using topology loss increases the continuity of the segmented vessels but results in poorer overlap-based quality, which translates into arteriole-venule classification mistakes. This issue was addressed by incorporating an orientation-score guided module using cake wavelets, a type of orientation sensitive filter which was previously used for junction point classification \cite{Abbasi-Sureshjani2016,Feliciano2016}. 

In conclusion, we novelly incorporate topology preserving losses and orientation-scores in a CNN-based vision model to obtain semantically accurate retinal AV segmentations. Both improving the cake-wavelet module design and extending the method to other modalities such as 3D computed tomography angiography are important directions for future work.


\bibliographystyle{spiebib} 

\end{document}